\documentclass[letterpaper, 10pt, conference]{ieeeconf}

\IEEEoverridecommandlockouts    

\usepackage{multirow}
\usepackage{lipsum}
\usepackage{amsmath}
\usepackage{amssymb}
\usepackage[ruled,vlined]{algorithm2e}
\usepackage{graphicx}
\graphicspath{{./Figures/}}
\usepackage[hidelinks]{hyperref}
\usepackage{booktabs}

\usepackage{tabularx}
\usepackage[table]{xcolor}
\usepackage{colortbl}

\definecolor{lightgreen}{RGB}{220,245,220}
\definecolor{lightred}{RGB}{255,220,220}

\newcommand{\improve}[1]{%
  \ifdim #1pt > 0pt
    \cellcolor{lightgreen}#1%
  \else
    \cellcolor{lightred}#1%
  \fi
}

\title{\LARGE \bf Time Series Foundation Models as Strong Baselines in Transportation Forecasting: A Large-Scale Benchmark Analysis}

\author{
\begin{tabular}{c}
Javier Yanes-Pulido$^{1}$, Filipe Rodrigues$^{1}$\\[0.5ex]
{\small $^{1}$Technical University of Denmark, Kgs. Lyngby, Denmark}\\
{\tt\small s243345@dtu.dk, rodr@dtu.dk}
\end{tabular}
}

\begin{document}
	
	\maketitle
	
	\begin{abstract}
		Accurate forecasting of transportation dynamics is essential for urban mobility and infrastructure planning. Although recent work has achieved strong performance with deep learning models, these methods typically require dataset-specific training, architecture design and hyper-parameter tuning. This paper evaluates whether general-purpose time-series foundation models can serve as forecasters for transportation tasks by benchmarking the zero-shot performance of the state-of-the-art model, Chronos-2, across ten real-world datasets covering highway traffic volume and flow, urban traffic speed, bike-sharing demand, and electric vehicle charging station data. Under a consistent evaluation protocol, we find that, even without any task-specific fine-tuning, Chronos-2 delivers state-of-the-art or competitive accuracy across most datasets, frequently outperforming classical statistical baselines and specialized deep learning architectures, particularly at longer horizons. Beyond point forecasting, we evaluate its native probabilistic outputs using prediction-interval coverage and sharpness, demonstrating that Chronos-2 also provides useful uncertainty quantification without dataset-specific training. In general, this study supports the adoption of time-series foundation models as a key baseline for transportation forecasting research.
	\end{abstract}
	
	\section{Introduction}
\label{sec:introduction}

The accurate prediction of traffic speeds, traffic volumes, passenger flows or electric vehicle (EV) charging infrastructure demand, among many other transportation dynamics, constitutes a fundamental pillar in the operational management of Intelligent Transportation Systems. Historically, researchers and urban planners have used classical parametric models for this task, such as the Autoregressive Integrated Moving Average (ARIMA). While these classical models offer high mathematical interpretability, they have many important limitations \cite{Reddy2025_ARIMA}. Consequently, the transportation forecasting discipline underwent a massive methodological shift toward deep learning (DL) architectures. With more high-resolution sensor data available and greater computational power, research has increasingly focused on complex neural network designs. Some examples include Spatial-Temporal Graph Convolutional Networks (STGCNs) \cite{Yu2018-Spatio-Temporal}, Diffusion Convolutional Recurrent Neural Networks (DCRNNs) \cite{li2018diffusionconvolutionalrecurrentneural}, Temporal Graph Convolutional Networks (T-GCNs) \cite{Zhao2020-T-GCN}, and 3-Dimensional Dynamic Graph Convolutional Networks (3DGCNs) \cite{Xia2021-3DGCN}.

At the same time, the broader field of artificial intelligence is experiencing a major paradigm shift driven by the emergence of Foundation Models (FMs). Initially developed in natural language processing through large-scale transformer architectures such as GPT-3 \cite{Brown2020_GPT-3} and BERT \cite{Devlin2018_BERT}, these models are trained on large heterogeneous corpora using self-supervised objectives to learn general representations. Unlike task-specific models trained from scratch, FMs capture broad statistical structure, enabling zero-shot generalization to unseen datasets. This paradigm has recently been extended to time series forecasting through Time Series Foundation Models (TS-FMs). Representative approaches include Time-LLM \cite{jin2024timellmtimeseriesforecasting},  PromptCast \cite{xue2023promptcastnewpromptbasedlearning}, TimesFM \cite{das2024googletimeseries}, Lag-Llama \cite{rasul2024lagllamafoundationmodelsprobabilistic}, and the Chronos family \cite{ansari2024chronoslearninglanguagetime, ansari2025chronos2univariateuniversalforecasting}. This shift represents a new operational paradigm, not merely another model added to the literature, and it is crucial to update transportation forecasting benchmarks accordingly.

Additionally, the integration of TS-FMs facilitates a methodological transition: moving from deterministic point estimates to probabilistic modeling. The traditional transportation forecasting literature has largely relied on deterministic error metrics such as Mean Absolute Error (MAE), Root Mean Squared Error (RMSE), and Mean Absolute Percentage Error (MAPE). However, deterministic point estimates have significant theoretical and practical limitations \cite{Hewamalage2023ForecastEvaluation}, and in transportation planning, understanding forecast variance and tail risks could be as important as the median expectation. TS-FMs like Chronos natively address these weaknesses by making it possible to forecast probabilistically. In this context, forecast quality can be evaluated through the dual lenses of calibration and sharpness, as formalized in \cite{Raftery2007-calibration-sharpness}.

The main contributions of this research are structured as follows:
\begin{itemize}
    \item We benchmark the state-of-the-art TS-FM Chronos-2 \cite{ansari2025chronos2univariateuniversalforecasting} across ten diverse real-world transportation datasets, spanning highway traffic speeds and volumes, urban traffic conditions, bike-sharing demand and flows, and EV station data.
    \item We demonstrate that zero-shot Chronos-2 achieves state-of-the-art or highly competitive performance on standard deterministic metrics (MAE, RMSE, MAPE), frequently outperforming both classical statistical methods and task-specific DL architectures, thus arguing for the inclusion of TS-FMs as standard baselines in future transportation forecasting research.
    \item Leveraging Chronos-2's native probabilistic outputs, we move beyond deterministic point estimates and assess forecast uncertainty through calibration (empirical coverage) and sharpness (interquantile range), thus enabling probabilistic evaluation without dataset-specific training or architectural modifications.
\end{itemize}

	\section{Methods}
\label{sec:designandimplementation}
This study builds on the methodologies from \cite{Rodrigues2023-on-the-importance-of-stationarity}, which compared statistical and DL models across nine diverse transportation datasets. We extend this work by including the newly introduced UrbanEV \cite{li2025urbanev}, which focuses on EV infrastructure, and applying the TS-FM paradigm via the Chronos-2 model on these ten out-of-domain datasets. The datasets vary in network structure, temporal resolution, and underlying processes (Table \ref{tab:datasets}), allowing assessment of how well models generalize across different urban mobility scenarios.

\begin{table*}[t]
\centering
\caption{Summary of Transportation Benchmark Datasets and Evaluated Horizons.}
\label{tab:datasets}

\setlength{\tabcolsep}{3pt}
\begin{tabular}{cccccccc}
\toprule
\textbf{Dataset} & \textbf{Type} & \textbf{Range [unit]} & \textbf{Time Span} & \textbf{Granularity} & \textbf{Train/Val/Test} & \textbf{Horizons Evaluated} & \textbf{Source} \\
\midrule
PeMSD7(M) & Traffic Speeds & 3--82.6 mph & 01/04/2016 -- 30/06/2016 & 5 mins & 34/5/5 days & 15, 30, 45 mins & \cite{Yu2018-Spatio-Temporal} \\
Urban1 & Traffic Speeds & 0.5--110 km/h & 01/04/2018 -- 30/04/2018 & 5 mins & 70/10/20 \% & 30, 45, 60 mins & \cite{lee2022ddpgcnmultigraphconvolutionalnetwork} \\
NYC Citi Bike & Pickups/Dropoffs & 0--108 trips & 01/04/2016 -- 30/06/2016 & 30 mins & 63/14/14 days & 1 to 12 steps & \cite{Ye_Sun_Du_Fu_Xiong_2021} \\
PeMSD4 & Traffic Volumes & 0--919 veh. & 01/01/2018 -- 28/02/2018 & 5 mins & 60/20/20 \% & 1 to 12 steps & \cite{Choi_Choi_Hwang_Park_2022} \\
SZ-taxi & Traffic Speeds & 0--86.4 km/h & 01/01/2015 -- 31/01/2015 & 15 mins & 80/-/20 \% & 15, 30, 45, 60 mins & \cite{Zhao2020-T-GCN} \\
METR-LA & Traffic Speeds & 0--70 mph & 01/03/2012 -- 30/06/2012 & 5 mins & 70/10/20 \% & 15, 30, 60 mins & \cite{li2018diffusionconvolutionalrecurrentneural} \\
PEMS-BAY & Traffic Speeds & 0--85.1 mph & 01/01/2017 -- 31/05/2017 & 5 mins & 70/10/20 \% & 15, 30, 60 mins & \cite{li2018diffusionconvolutionalrecurrentneural} \\
NYC Bike Flow & In/Out Flows & 0--434 trips & 01/07/2017 -- 30/09/2017 & 1 hour & 80/10/10 \% & 1, 2, 3 hours & \cite{Xia2021-3DGCN} \\
Seattle Loop & Traffic Speeds & 0.8--80.7 mph & 01/11/2015 -- 31/12/2015 & 5 mins & 56/-/5 days & 5 mins & \cite{Yang2021-real-time-spatiotemporal} \\

\multirow{3}{*}{UrbanEV} 
& Station Occupancy & 0--373 piles
& \multirow{3}{*}{01/09/2022 -- 28/02/2023}
& \multirow{3}{*}{1 hour}
& \multirow{3}{*}{80/10/10 \%}
& \multirow{3}{*}{3, 6, 9, 12 hours}
& \multirow{3}{*}{\cite{li2025urbanev}} \\

& Session Duration & 0--207.6 min & & & & & \\

& Charging Volume & 0--16{,}732.5 kWh & & & & & \\
\bottomrule
\end{tabular}
\end{table*}

This study employs the state-of-the-art Chronos-2 model, which extends earlier univariate TS-FMs to universal forecasting across univariate, multivariate, and covariate-informed settings \cite{ansari2025chronos2univariateuniversalforecasting}. The architecture is encoder-only and transformer-based, derived from the Text-to-Text Transfer Transformer (T5) (a widely adopted foundational large language model) \cite{Raffel2019_T5}. Its central innovation is a \textit{group attention} mechanism that alternates between temporal attention within each series and cross-series attention within predefined groups (e.g., sensors or EV charging stations). This design enables cross-learning among series, which naturally exploits spatial correlations by sharing information within a batch of observations, without requiring predefined adjacency matrices. This differentiates the approach from graph neural network-based methods, which typically rely on fixed adjacency matrices to model spatial structure \cite{Yu2018-Spatio-Temporal, li2018diffusionconvolutionalrecurrentneural, Zhao2020-T-GCN}. Another key advantage of TS-FMs like Chronos-2 is their innate ability to go beyond single-point predictions and provide full probabilistic forecasts. Chronos-2 natively outputs a rich representation of the predictive distribution. In a single forward pass, the model directly predicts a fixed set of 21 quantiles (ranging from 0.01 to 0.99) for the specified forecasting horizon.

We began by choosing all nine datasets from \cite{Rodrigues2023-on-the-importance-of-stationarity} and establishing an evaluation framework that ensured we could strictly reproduce prior results. We subsequently integrated the more recent UrbanEV dataset, so that the benchmark analysis also covers EV-charging-related tasks. To maintain direct comparability, the evaluation utilizes the same procedure established in prior literature. The models were evaluated using a deterministic sliding window procedure across the designated test sets \cite{Yu2018-Spatio-Temporal, lee2022ddpgcnmultigraphconvolutionalnetwork, Ye_Sun_Du_Fu_Xiong_2021, Choi_Choi_Hwang_Park_2022, Zhao2020-T-GCN, li2018diffusionconvolutionalrecurrentneural, Xia2021-3DGCN, Yang2021-real-time-spatiotemporal, li2025urbanev}. The test set size corresponds to a different specific percentage of the whole dataset for each respective dataset (Table \ref{tab:datasets}). The evaluation window rolls forward exactly one step at a time, generating predictions for the predefined forecasting horizons with all splits strictly chronological. For our experimental setup, inference was executed on a single GPU, and we utilized the \href{https://huggingface.co/amazon/chronos-2}{\texttt{amazon/chronos-2}} model (120M parameters) via the \href{https://github.com/amazon-science/chronos-forecasting}{\texttt{chronos-forecasting}} library (v2.2.2), with cross-learning enabled across all experiments.

A critical parameter in utilizing TS-FMs is defining the context window, i.e. the amount of historical data provided to the model to condition its zero-shot inference. We used as context for this model the past week of historical data. We chose this duration to balance sufficient historical visibility (capturing the recurrent weekly seasonality identified by the historical average (HA) baseline) against inference performance and computational cost. Because the datasets vary in granularity (5-minute to 1-hour intervals), preserving a one-week historical context required a dataset-specific number of context tokens that was determined utilizing the formula:
\begin{equation}
    \text{Context tokens} = \frac{\text{Minutes per week}}{\text{Granularity}}.
\end{equation}
For example, the PeMSD4 dataset (5-minute granularity) required an input sequence of 2016 context tokens, whereas the UrbanEV dataset (60-minute granularity) required only 168 context tokens.

In datasets containing more than one target variable (i.e., the bike datasets), we utilized the alternate variate as a known covariate. For example, in the NYC Citi Bike dataset, we predicted station dropoffs by also providing the corresponding pickups at previous time steps as in-context covariates, and then performed the reverse operation to predict pickups, averaging the final results.

Since Chronos-2 can produce distributional outputs, our evaluation is designed to (i) remain comparable to prior work and (ii) leverage the model's native uncertainty estimates. To maintain comparability with prior work, we report MAE, RMSE, and MAPE using the 0.5 quantile (median) as the point forecast. For MAPE, ground-truth zero values are excluded via masking for all datasets except UrbanEV, where values $\leq 0.02$ are instead adjusted using a small $\varepsilon$ offset following the original benchmark protocol \cite{li2025urbanev}. Additionally, we perform a probabilistic evaluation using the 0.1 and 0.9 quantiles to form an 80\% prediction interval, measuring calibration as empirical coverage (target: 80\%) and sharpness as the interquantile range (IQR) between those bounds.

	\section{Results}
\label{sec:results}

We divide our evaluation into two components: a deterministic point-estimate comparison against established baselines, and a probabilistic evaluation of Chronos-2's uncertainty quantification. All results are reproducible via the project's GitHub repository (\url{https://github.com/fmpr/mobility-baselines}). Values reported directly from the original publications are marked with an asterisk (*), and all evaluations follow the protocols and source code provided by the original authors.

\textbf{Deterministic point-estimate comparison.} Tables \ref{tab:PEMSD7(M)}–\ref{tab:UrbanEV_Volume} report deterministic forecasting performance across the ten datasets. The zero-shot TS-FM (Chronos-2) achieves state-of-the-art or highly competitive results, frequently surpassing statistical baselines and advanced DL models. Performance remains stable over longer horizons, contrasting with traditional autoregressive approaches where error accumulation typically degrades accuracy. Chronos-2 exhibits reduced sensitivity to horizon length, maintaining competitive accuracy as the forecast window expands. Table \ref{tab:summary_improvements} summarizes improvements in MAE, MAPE, and RMSE relative to the statistical baseline (HA) and the strongest DL competitor, showing substantial error reductions and top rankings on most datasets, with METR-LA as an exception potentially attributable to dataset-specific dynamics. Tables \ref{tab:UrbanEV_Duration} and \ref{tab:UrbanEV_Volume} include only the Last Observation (LO) baseline, which uses the last observation as the forecast for the next time step, and the best DL model (TimeXer) to focus on the most relevant comparisons.

\begin{table}[t!]
    \centering
    \caption{RESULTS FOR PEMSD7(M) - TRAFFIC SPEEDS IN CALIFORNIA.}
    \label{tab:PEMSD7(M)}
    \begin{tabular}{c|c|c|c}
     & MAE & MAPE & RMSE \\
    Model & 15/ 30/ 45 min & 15/ 30/ 45 min & 15/ 30/ 45 min \\
    \hline
    LSVR* & 2.50/ 3.63/ 4.54 & 5.81/ 8.88/ 11.50 & 4.55/ 6.67/ 8.28 \\
    FNN* & 2.74/ 4.02/ 5.04 & 6.38/ 9.72/ 12.38 & 4.75/ 6.98/ 8.58 \\
    FC-LSTM* & 3.57/ 3.94/ 4.16 & 8.60/ 9.55/ 10.10 & 6.20/ 7.03/ 7.51 \\
    GCGRU* & 2.37/ 3.31/ 4.01 & 5.54/ 8.06/ 9.99 & 4.21/ 5.96/ 7.13 \\
    STGCN(Cheb) & \textbf{2.25}/ 3.03/ 3.57 & 5.26/ 7.33/ 8.69 & 4.04/ 5.70/ 6.77 \\
    STGCN(1st) & 2.26/ 3.09/ 3.79 & \textbf{5.24}/ 7.39/ 9.12 & 4.07/ 5.77/ 7.03 \\
    \hline
    HA & 3.90 & 10.14 & 7.09 \\
    HA+LR & 2.48/ 3.13/ 3.45 & 5.81/ 7.65/ 8.57 & 4.22/ 5.50/ 6.10 \\
    \hline
    Chronos-2 & \textbf{2.25}/ \textbf{2.67}/ \textbf{2.90} & 5.33/ \textbf{6.61}/ \textbf{7.31} & \textbf{4.10}/ \textbf{4.94}/ \textbf{5.37} \\
    \hline
    \end{tabular}
\end{table}

\begin{table}[t!]
    \centering
    \caption{RESULTS FOR URBAN1 - TRAFFIC SPEEDS IN SOUTH KOREA.}
    \begin{tabular}{c|c|c|c}
     & MAE & MAPE & RMSE \\
    Model & 30/ 45/ 60 min & 30/ 45/ 60 min & 30/ 45/ 60 min \\
    \hline
    VAR* & 5.06/ 4.99/ 4.97 & 23.10/ 22.82/ 22.73 & 7.04/ 6.92/ 6.88 \\
    LSVR* & 3.82/ 3.89/ 3.93 & 15.35/ 17.99/ 17.39 & 5.64/ 5.74/ 5.84 \\
    ARIMA* & 3.49/ 3.79/ 4.04 & 15.40/ 16.85/ 18.09 & 5.28/ 5.65/ 5.94 \\
    FC-LSTM* & 3.91/ 3.92/ 3.92 & 17.29/ 17.32/ 17.31 & 6.38/ 6.39/ 6.39 \\
    DCRNN* & 3.17/ 3.46/ 3.73 & 13.52/ 14.83/ 15.95 & 4.94/ 5.30/ 5.61 \\
    STGCN* & 3.07/ 3.42/ 3.80 & 14.38/ 16.72/ 19.37 & 4.57/ 4.83/ 5.04 \\
    DDP-GCN & 3.00/ 3.00/ 2.99 & 13.57/ 13.56/ 13.51 & 4.45/ 4.45/ 4.47 \\
    \hline
    HA & 3.18 & 14.19 & 4.79 \\
    HA+LR & 3.04/ 3.10/ 3.13 & 13.39/ 13.73/ 13.87 & 4.60/ 4.67/ 4.71 \\
    \hline
    Chronos-2 & \textbf{2.71}/ \textbf{2.78}/ \textbf{2.83} & \textbf{11.97}/ \textbf{12.38}/ \textbf{12.71} & \textbf{4.03}/ \textbf{4.12}/ \textbf{4.17} \\
    \hline
    \end{tabular}
\end{table}

\begin{table}[t!]
    \centering
    \caption{RESULTS FOR NYC CITI BIKE - PICKUPS AND DROPOFFS IN NYC.}
    \begin{tabular}{c|c|c}
    Model & MAE & RMSE \\
    \hline
    XGBoost* & 2.469 & 4.050 \\
    FC-LSTM* & 2.303 & 3.814 \\
    DCRNN* & 1.895 & 3.209 \\
    STGCN* & 2.761 & 3.604 \\
    STG2Seq* & 2.498 & 3.984 \\
    Graph WaveNet* & 1.991 & 3.294 \\
    CCRNN & 1.740 & 2.838 \\
    \hline
    HA & 1.726 & 2.871 \\
    HA+LR & 1.738 & 2.758 \\
    \hline
    Chronos-2 & \textbf{1.574} & \textbf{2.239} \\
    \hline
    \end{tabular}
\end{table}

\begin{table}[t!]
    \centering
    \caption{RESULTS FOR PEMSD4 - TRAFFIC VOLUMES IN CALIFORNIA.}
    \begin{tabular}{c|c|c|c}
    Model & MAE & MAPE & RMSE \\
    \hline
    ARIMA* & 33.73 & 24.18 & 48.80 \\
    VAR* & 24.54 & 17.24 & 38.61 \\
    FC-LSTM* & 26.77 & 18.23 & 40.65 \\
    TCN* & 23.22 & 15.59 & 37.26 \\
    GRU-ED* & 23.68 & 16.44 & 39.27 \\
    DSANet* & 22.79 & 16.03 & 35.77 \\
    STGCN* & 21.16 & 13.83 & 34.89 \\
    DCRNN* & 21.22 & 14.17 & 33.44 \\
    GraphWaveNet* & 24.89 & 17.29 & 39.66 \\
    ASTGCN(r)* & 22.93 & 16.56 & 35.22 \\
    MSTGCN* & 23.96 & 14.33 & 37.21 \\
    STG2Seq* & 25.20 & 18.77 & 38.48 \\
    LSGCN* & 21.53 & 13.18 & 33.86 \\
    STSGCN* & 21.19 & 13.90 & 33.65 \\
    AGCRN* & 19.83 & 12.97 & 32.26 \\
    STFGNN* & 20.48 & 16.77 & 32.51 \\
    STGODE* & 20.84 & 13.77 & 32.82 \\
    Z-GCNETs* & 19.50 & 12.78 & 31.61 \\
    STG-NCDE & 19.21 & \textbf{12.76} & 31.09 \\
    \hline
    HA & 26.26 & 17.07 & 42.87 \\
    HA+LR & 20.03 & 13.39 & 32.73 \\
    \hline
    Chronos-2 & \textbf{18.83} & 13.57 & \textbf{28.30} \\
    \hline
    \end{tabular}
\end{table}

\begin{table}[t!]
    \centering
    \caption{RESULTS FOR SZ-TAXI - TRAFFIC SPEEDS IN SHENZHEN FROM TAXI TRAJECTORIES.}
    \begin{tabular}{c|c|c}
     & MAE & RMSE \\
    Model & 15/ 30/ 45/ 60 min & 15/ 30/ 45/ 60 min \\
    \hline
    T-GCN & 4.517/ 4.572/ 4.621/ 4.671 & 5.997/ 6.034/ 6.064/ 6.100 \\
    \hline
    HA & 4.630 & 6.463 \\
    HA+LR & 3.464/ 3.507/ 3.534/ 3.554 & 4.998/ 5.057/ 5.091/ 5.115 \\
    \hline
    Chronos-2 & \textbf{2.463}/ \textbf{2.468}/ \textbf{2.470}/ \textbf{2.472} & \textbf{3.912}/ \textbf{3.919}/ \textbf{3.925}/ \textbf{3.929} \\
    \hline
    \end{tabular}
\end{table}

\begin{table}[t!]
    \centering
    \caption{RESULTS FOR METR-LA - TRAFFIC SPEEDS IN LA.}
    \begin{tabular}{c|c|c|c}
     & MAE & MAPE & RMSE \\
    Model & 15/ 30/ 60 min & 15/ 30/ 60 min & 15/ 30/ 60 min \\
    \hline
    ARIMA* & 3.99/ 5.15/ 6.90 & 9.6/ 12.7/ 17.4 & 8.21/ 10.45/ 13.23 \\
    VAR* & 4.42/ 5.41/ 6.52 & 10.2/ 12.7/ 15.8 & 7.89/ 9.13/ 10.11 \\
    SVR* & 3.99/ 5.05/ 6.72 & 9.3/ 12.1/ 16.7 & 8.45/ 10.87/ 13.76 \\
    FNN* & 3.99/ 4.23/ 4.49 & 9.9/ 12.9/ 14.0 & 7.94/ 8.17/ 8.69 \\
    FC-LSTM* & 3.44/ 3.77/ 4.37 & 9.6/ 10.9/ 13.2 & 6.30/ 7.23/ 8.69 \\
    DCRNN & \textbf{2.77}/ \textbf{3.15}/ \textbf{3.60} & \textbf{7.3}/ \textbf{8.8}/ \textbf{10.5} & \textbf{5.38}/ \textbf{6.45}/ 7.59 \\
    \hline
    HA & 4.19 & 13.0 & 7.84 \\
    HA+LR & 3.28/ 3.68/ 4.02 & 8.8/ 10.4/ 11.9 & 5.71/ 6.60/ \textbf{7.32} \\
    \hline
    Chronos-2 & 3.31/ 3.95/ 4.72 & 8.16/ 9.94/ 12.03 & 5.90/ 6.89/ 7.98 \\
    \hline
    \end{tabular}
\end{table}

\begin{table}[t!]
    \centering
    \caption{RESULTS FOR PEMS-BAY - TRAFFIC SPEEDS IN CALIFORNIA.}
    \begin{tabular}{c|c|c|c}
     & MAE & MAPE & RMSE \\
    Model & 15/ 30/ 60 min & 15/ 30/ 60 min & 15/ 30/ 60 min \\
    \hline
    ARIMA* & 1.62/ 2.33/ 3.38 & 3.5/ 5.4/ 8.3 & 3.30/ 4.76/ 6.50 \\
    VAR* & 1.74/ 2.32/ 2.93 & 3.6/ 5.0/ 6.5 & 3.16/ 4.25/ 5.44 \\
    SVR* & 1.85/ 2.48/ 3.28 & 3.8/ 5.5/ 8.0 & 3.59/ 5.18/ 7.08 \\
    FNN* & 2.20/ 2.30/ 2.46 & 5.2/ 5.4/ 5.9 & 4.42/ 4.63/ 4.98 \\
    FC-LSTM* & 2.05/ 2.20/ 2.37 & 4.8/ 5.2/ 5.7 & 4.19/ 4.55/ 4.96 \\
    DCRNN & \textbf{1.38}/ 1.74/ 2.07 & \textbf{2.9}/ \textbf{3.9}/ 4.9 & 2.95/ 3.97/ 4.74 \\
    \hline
    HA & 2.58 & 6.1 & 5.04 \\
    HA+LR & 1.54/ 1.91/ 2.22 & 3.2/ 4.3/ 5.1 & 2.93/ 3.83/ 4.45 \\
    \hline
    Chronos-2 & 1.46/ \textbf{1.71}/ \textbf{1.95} & 3.2/ \textbf{3.9}/ \textbf{4.7} & \textbf{2.81}/ \textbf{3.34}/ \textbf{3.78} \\
    \hline
    \end{tabular}
\end{table}

\begin{table}[t!]
    \centering
    \caption{RESULTS FOR NYC BIKE IN- AND OUT-FLOWS.}
    \begin{tabular}{c|c|c}
     & MAE & RMSE \\
    Model & 1h/ 2h/ 3h & 1h/ 2h/ 3h \\
    \hline
    ARIMA* & 10.41/ 11.84/ 13.00 & 19.14/ 21.76/ 23.90 \\
    STGCN* & 6.49/ 7.06/ 7.94 & 11.73/ 12.93/ 15.37 \\
    DCRNN* & 5.88/ 6.19/ 7.72 & 9.85/ 10.39/ 12.37 \\
    STGNN* & 5.79/ 6.00/ 7.56 & 9.80/ 9.98/ 11.91 \\
    MVGCN* & 5.65/ 7.72/ 8.00 & 9.64/ 13.53/ 13.93 \\
    3DGCN & 4.81/ 5.61/ 6.99 & 7.76/ 9.49/ 11.74 \\
    \hline
    HA & 5.97 & 11.04 \\
    HA+LR & 5.10/ 5.45/ 5.56 & 8.72/ 9.69/ 10.04 \\
    \hline
    Chronos-2 & \textbf{4.78}/ \textbf{4.79}/ \textbf{4.76} & \textbf{6.89}/ \textbf{6.90}/ \textbf{6.86} \\
    \hline
    \end{tabular}
\end{table}

\begin{table}[t!]
\centering
    \caption{RESULTS FOR SEATTLE TRAFFIC SPEED DATA.}
    \begin{tabular}{c|c|c}
     & MAPE & RMSE \\
    \hline
    BTMF* & 7.70 & 4.59 \\
    TRMF-GRMF* & 8.20 & 4.83 \\
    TRMF-ALS* & 8.36 & 4.96 \\
    Linear-LSTM-ReMF* & 8.01 & 4.59 \\
    BiLSTM-GL-ReMF & 7.83 & 4.50 \\
    LSTM-ReMF & 7.64 & 4.42 \\
    LSTM-GL-ReMF & 7.64 & 4.43 \\
    \hline
    HA & 12.5 & 9.82 \\
    HA+LR & 5.5 & 3.95 \\
    \hline
    Chronos-2 & \textbf{5.36} & \textbf{3.74} \\
    \hline
    \end{tabular}
\end{table}

\begin{table*}[t!]
    \centering
    \caption{RESULTS FOR URBANEV - EV STATION OCCUPANCY IN SHENZHEN.}
    \label{tab:UrbanEV_Occ}
    \setlength{\tabcolsep}{2pt}
    \begin{tabular}{c|c|c|c|c}
     & RMSE ($\times 10^{-2}$) & MAPE (\%) & RAE ($\times 10^{-2}$) & MAE ($\times 10^{-2}$) \\
    Model & 3/ 6/ 9/ 12 h (Avg) & 3/ 6/ 9/ 12 h (Avg) & 3/ 6/ 9/ 12 h (Avg) & 3/ 6/ 9/ 12 h (Avg) \\
    \hline
    LO & 8.32/ 10.6/ 12.38/ 13.06 (11.09) & 24.83/ 37.55/ 48.57/ 53.89 (41.21) & 33.31/ 49.02/ 61.69/ 67.25 (52.82) & 4.04/ 5.97/ 7.54/ 8.22 (6.44) \\
    AR & 10.39/ 10.73/ 11.0/ 11.04 (10.79) & 57.64/ 58.85/ 59.72/ 59.91 (59.03) & 59.08/ 60.78/ 61.9/ 62.02 (60.94) & 7.18/ 7.4/ 7.57/ 7.58 (7.43) \\
    ARIMA & 10.32/ 10.94/ 11.48/ 11.67 (11.1) & 52.18/ 54.72/ 56.97/ 57.88 (55.44) & 56.55/ 60.09/ 62.98/ 63.97 (60.89) & 6.86/ 7.31/ 7.69/ 7.81 (7.42) \\
    FCNN & 7.86/ 8.95/ 9.16/ 8.23 (8.55) & 27.36/ 37.08/ 39.46/ 35.21 (34.78) & 35.02/ 45.0/ 46.93/ 41.07 (42.01) & 4.25/ 5.49/ 5.74/ 5.03 (5.13) \\
    LSTM & 8.15/ 9.43/ 9.58/ 8.42 (8.9) & 28.03/ 39.45/ 40.71/ 35.69 (35.97) & 36.9/ 48.51/ 49.85/ 42.36 (44.4) & 4.48/ 5.93/ 6.11/ 5.19 (5.43) \\
    GCN & 7.03/ 8.13/ 8.97/ 8.31 (8.11) & 30.17/ 38.77/ 44.57/ 40.59 (38.53) & 35.52/ 43.87/ 49.42/ 45.23 (43.51) & 4.34/ 5.38/ 6.08/ 5.56 (5.34) \\
    GCNLSTM & 6.98/ 8.41/ 9.05/ 8.45 (8.22) & 30.1/ 42.06/ 46.83/ 43.42 (40.6) & 35.38/ 45.91/ 50.1/ 46.25 (44.41) & 4.31/ 5.62/ 6.16/ 5.69 (5.45) \\
    ASTGCN & 7.75/ 8.92/ 9.12/ 8.22 (8.5) & 28.06/ 37.22/ 39.22/ 35.11 (34.9) & 35.39/ 44.73/ 46.89/ 41.52 (42.13) & 4.31/ 5.46/ 5.74/ 5.08 (5.15) \\
    TimeXer & 7.26/ 8.16/ 8.64/ 8.1 (8.04) & 23.87/ 31.05/ 35.5/ 32.82 (30.81) & 31.31/ 38.95/ 43.21/ 40.22 (38.42) & 3.83/ 4.74/ 5.24/ 4.88 (4.67) \\
    \hline
    Chronos-2 & \textbf{4.57}/ \textbf{4.59}/ \textbf{4.59}/ \textbf{4.57} (\textbf{4.58}) & \textbf{12.17}/ \textbf{12.16}/ \textbf{12.13}/ \textbf{12.11} (\textbf{12.14}) & \textbf{15.77}/ \textbf{15.74}/ \textbf{15.65}/ \textbf{15.59} (\textbf{15.69}) & \textbf{1.90}/ \textbf{1.91}/ \textbf{1.90}/ \textbf{1.89} (\textbf{1.90}) \\
    \hline
    \end{tabular}
\end{table*}

\begin{table*}[t!]
    \centering
    \caption{RESULTS FOR URBANEV - EV STATION CHARGING SESSION DURATION IN SHENZHEN.}
    \label{tab:UrbanEV_Duration}
    \begin{tabular}{c|c|c|c|c}
     & RMSE & MAPE (\%) & RAE & MAE \\
    Model & 3/ 6/ 9/ 12 h (Avg) & 3/ 6/ 9/ 12 h (Avg) & 3/ 6/ 9/ 12 h (Avg) & 3/ 6/ 9/ 12 h (Avg) \\
    \hline
    LO & 5.80 / 8.53/ 10.66/ 11.45 (9.11) & 1.24/ 2.23/ 3.20/ 3.78 (2.61) & 0.25/ 0.39/ 0.51/ 0.56 (0.43) & 2.63/ 4.22/ 5.57/ 6.12 (4.64) \\
    TimeXer & 4.58/ 5.23/ 5.75/ 4.94 (5.13) & 1.47/ 2.10/ 2.62/ 2.40 (2.15) & 0.21/ 0.27/ 0.31/ 0.27 (0.26) & 2.25/ 2.82/ 3.18/ 2.81 (2.76) \\
    \hline
    Chronos-2 & \textbf{1.55}/ \textbf{1.55}/ \textbf{1.55}/ \textbf{1.55} (\textbf{1.55}) & \textbf{0.45}/ \textbf{0.45}/ \textbf{0.45}/ \textbf{0.45} (\textbf{0.45}) & \textbf{0.08}/ \textbf{0.08}/ \textbf{0.08}/ \textbf{0.08} (\textbf{0.08}) & \textbf{0.82}/ \textbf{0.82}/ \textbf{0.82}/ \textbf{0.81} (\textbf{0.82}) \\
    \hline
    \end{tabular}
\end{table*}

\begin{table*}[t!]
    \centering
    \scriptsize
    \caption{RESULTS FOR URBANEV - EV STATION CHARGING VOLUME IN SHENZHEN.}
    \label{tab:UrbanEV_Volume}
    \begin{tabular}{c|c|c|c|c}
     & RMSE & MAPE (\%) & RAE & MAE \\
    Model & 3/ 6/ 9/ 12 h (Avg) & 3/ 6/ 9/ 12 h (Avg) & 3/ 6/ 9/ 12 h (Avg) & 3/ 6/ 9/ 12 h (Avg) \\
    \hline
    LO & 489.15/ 556.71/ 665.07/ 646.01 (589.24) & 18.95/ 29.77/ 46.44/ 54.76 (37.48) & 0.32/ 0.39/ 0.48/ 0.48 (0.42) & 102.00/ 127.18/ 157.37/ 157.59 (136.04) \\
    TimeXer & 353.63/ 376.01/ 370.44/ 299.58 (349.92) & 25.91/ 32.50/ 42.66/ 35.53 (34.15) & 0.25/ 0.28/ 0.29/ 0.24 (0.27) & 80.96/ 89.81/ 93.86/ 78.10 (85.68) \\
    \hline
    Chronos-2 & \textbf{140.85}/ \textbf{140.95}/ \textbf{140.77}/ \textbf{140.87} (\textbf{140.86}) & \textbf{8.87}/ \textbf{8.96}/ \textbf{9.07}/ \textbf{9.20} (\textbf{9.03}) & \textbf{0.10}/ \textbf{0.10}/ \textbf{0.10}/ \textbf{0.10} (\textbf{0.10}) & \textbf{31.65}/ \textbf{31.66}/ \textbf{31.62}/ \textbf{31.60} (\textbf{31.63}) \\
    \hline
    \end{tabular}
\end{table*}

\begin{table*}[t!]
\centering
\caption{Summary of Chronos-2 performance across benchmarks. 
Values denote the average percentage reduction in forecasting error relative to the statistical baseline (HA) and the strongest deep learning (DL) competitor. When multiple horizons are reported without averaging, the longest one is used. Rank denotes the rounded average rank across available metrics.}
\label{tab:summary_improvements}
\scriptsize
\setlength{\tabcolsep}{2pt}
\renewcommand{\arraystretch}{0.86}
\begin{tabularx}{\textwidth}{l c *{6}{>{\centering\arraybackslash}X}}
\toprule
\multirow{2}{*}{\textbf{Dataset}} 
& \multirow{2}{*}{\textbf{Rank}} 
& \multicolumn{3}{c}{\textbf{Improvement vs.\ HA (\%)}} 
& \multicolumn{3}{c}{\textbf{Improvement vs.\ Best DL (\%)}} \\
\cmidrule(lr){3-5} \cmidrule(lr){6-8}
& 
& \textbf{MAE} & \textbf{MAPE} & \textbf{RMSE}
& \textbf{MAE} & \textbf{MAPE} & \textbf{RMSE} \\
\midrule

PeMSD7(M) & 1 
& \improve{+25.6} & \improve{+27.9} & \improve{+24.3}
& \improve{+18.8} & \improve{+15.9} & \improve{+12.7} \\

Urban1 & 1 
& \improve{+11.0} & \improve{+10.4} & \improve{+12.9}
& \improve{+5.4} & \improve{+5.9} & \improve{+6.7} \\

NYC Citi Bike & 1 
& \improve{+8.8} & - & \improve{+22.0}
& \improve{+9.5} & - & \improve{+21.1} \\

PeMSD4 & 1 
& \improve{+28.3} & \improve{+20.5} & \improve{+34.0}
& \improve{+2.0} & \improve{-6.3} & \improve{+9.0} \\

SZ-taxi & 1 
& \improve{+46.6} & - & \improve{+39.2}
& \improve{+47.1} & - & \improve{+35.6} \\

METR-LA & 3 
& \improve{-12.6} & \improve{+7.5} & \improve{-1.8}
& \improve{-31.1} & \improve{-14.6} & \improve{-5.1} \\

PEMS-BAY & 1 
& \improve{+24.4} & \improve{+23.0} & \improve{+25.0}
& \improve{+5.8} & \improve{+4.1} & \improve{+20.3} \\

NYC Bike Flow & 1 
& \improve{+20.3} & - & \improve{+37.9}
& \improve{+31.9} & - & \improve{+41.6} \\

Seattle Loop & 1 
& - & \improve{+57.1} & \improve{+61.9}
& - & \improve{+29.8} & \improve{+15.4} \\

UrbanEV Occupancy\textsuperscript{*} & 1 
& \improve{+70.5} & \improve{+70.5} & \improve{+58.7}
& \improve{+63.0} & \improve{+60.6} & \improve{+43.0} \\

UrbanEV Duration\textsuperscript{*} & 1 
& \improve{+82.3} & \improve{+82.8} & \improve{+83.0}
& \improve{+70.3} & \improve{+79.1} & \improve{+69.8} \\

UrbanEV Volume\textsuperscript{*} & 1 
& \improve{+76.7} & \improve{+75.9} & \improve{+76.1}
& \improve{+63.1} & \improve{+73.6} & \improve{+59.7} \\

\bottomrule
\multicolumn{8}{l}{\footnotesize \textsuperscript{*}For UrbanEV data, Last Observation (LO) is used as the statistical baseline instead of HA to remain consistent with the original paper \cite{li2025urbanev}.}
\end{tabularx}
\end{table*}

Overall, Chronos-2 performs particularly well on datasets with long-horizon forecasts, where its zero-shot generalization mitigates error accumulation. Performance improvements seem comparatively less pronounced on datasets dominated by complex spatial interactions (e.g. METR-LA), suggesting potential benefits from incorporating explicit adjacency matrices. The results indicate that TS-FMs provide a robust alternative to dataset-specific architectures, especially for applications requiring stable long-horizon forecasting and minimal task-specific tuning. Moreover, by providing strong zero-shot performance, TS-FMs present themselves as excellent candidates for applications for which the amount of available historical data is small (e.g., a new mobility service or new transit lines).

\textbf{Probabilistic evaluation.} To address limitations of deterministic point estimates, we evaluate the predictive distributions of Chronos-2 in terms of calibration (empirical coverage) and sharpness (IQR). Table \ref{tab:probabilistic_results} is intended to establish a benchmark for these metrics across the evaluated datasets. Results indicate that Chronos-2 delivers strong zero-shot uncertainty quantification: for an 80\% nominal interval, empirical coverage is generally close to the target; under-coverage on SZ-taxi and NYC Citi Bike suggests that datasets with frequent zero observations are harder to calibrate. Absolute IQR magnitudes vary across datasets due to differences in target variable ranges (Table \ref{tab:datasets}).

Table \ref{tab:UrbanEV_prob} reports results for UrbanEV Occupancy, selected as a representative case, where we equipped DL models with quantile losses, and statistical baselines were required to derive probabilistic intervals via analytical quantile estimation. Chronos-2 maintains high sharpness with narrow, informative intervals, whereas other models commonly over-cover at the expense of wider IQRs. Chronos-2 slightly under-covers but has tighter intervals, reflecting greater predictive confidence. This trade-off behavior aligns with the standard paradigm in probabilistic forecasting \cite{Raftery2007-calibration-sharpness}, which balances sharp predictive intervals with well-calibrated coverage.

\begin{table*}[t!]
    \centering
    \caption{RESULTS FOR URBANEV (PROBABILISTIC METRICS) - EV STATION OCCUPANCY IN SHENZHEN.}
    \label{tab:UrbanEV_prob}
    \begin{tabular}{c|c|c}
     & Coverage (\% , target 80\%) & IQR (\(\times 10^{-2}\)) \\
    Model & 3/ 6/ 9/ 12 h (Avg) & 3/ 6/ 9/ 12 h (Avg) \\
    \hline
    AR & 84.84/ 84.76/ 84.82/ 84.89 (84.83) & 26.29/ 27.15/ 27.84/ 27.96 (27.31) \\
    ARIMA & 86.11/ 85.90/ 85.65/ 85.60 (85.82) & 26.22/ 27.83/ 29.21/ 29.69 (28.24) \\
    FCNN & 84.20/ 84.60/ 84.49/ 85.04 (84.58) & 18.48/ 23.21/ 24.69/ 22.78 (22.29) \\
    LSTM & 83.18/ 83.75/ 84.85/ 84.72 (84.12) & 16.15/ 21.30/ 22.76/ 19.48 (19.92) \\
    GCN & 74.07/ 72.85/ 72.75/ 75.61 (73.82) & 12.12/ 14.16/ 15.98/ 16.19 (14.61) \\
    GCNLSTM & 77.66/ 71.16/ 73.26/ 75.74 (74.46) & 13.35/ 14.20/ 16.09/ 16.10 (14.94) \\
    ASTGCN & 82.90/ 83.10/ 85.63/ 85.47 (84.28) & 17.06/ 20.78/ 22.68/ 20.02 (20.14) \\
    \hline
    Chronos-2 & \textbf{78.27}/ \textbf{78.23}/ \textbf{78.21}/ \textbf{78.24} (\textbf{78.24}) & \textbf{5.60}/ \textbf{5.60}/ \textbf{5.58}/ \textbf{5.56} (\textbf{5.59}) \\
    \hline
    \end{tabular}
\end{table*}

\begin{table}[t!]
\centering
\caption{PROBABILISTIC EVALUATION OF CHRONOS-2: CALIBRATION (EMPIRICAL COVERAGE) AND SHARPNESS (IQR), AVERAGED ACROSS HORIZONS FOR AN 80\% NOMINAL INTERVAL.}
\label{tab:probabilistic_results}
\begin{tabular}{c|c|c}
Dataset & Coverage (\%) & IQR \\
\hline
PeMSD7(M) & 77.69 & 7.54 \\
Urban1 & 79.79 & 8.65 \\
NYC Citi Bike & 63.87 & 4.90 \\
PeMSD4 & 77.12 & 57.93 \\
SZ-taxi & 63.42 & 7.41 \\
METR-LA & 71.46 & 12.26 \\
PEMS-BAY & 77.94 & 4.77 \\
NYC Bike Flow & 77.15 & 16.38 \\
Seattle Loop & 79.87 & 7.41 \\
UrbanEV Occupancy & 78.24 & 5.59 $\times 10^{-2}$ \\
UrbanEV Duration & 78.82 & 2.61 \\
UrbanEV Volume & 78.51 & 95.21 \\
\hline
\end{tabular}
\end{table}

	\section{Discussion}
\label{sec:discussion}

\textbf{Evaluating a new paradigm.} This study evaluates whether a generalist, pre-trained TS-FM can serve as a competitive forecaster across diverse transportation tasks without task-specific architecture engineering, making a case for TS-FMs as a new benchmarking standard in the field.

\textbf{Performance and simplicity of usage.} The empirical results confirm that using TS-FMs in the transportation domain is not only possible, but highly effective. Beyond raw performance, Chronos-2 is notably simple to use. Traditional DL architectures typically require explicit spatial adjacency matrices, exhaustive hyperparameter tuning, and significant computational resources for training on each new dataset/task. In contrast, Chronos-2 makes local inference on standard hardware like a laptop feasible.

\textbf{Probabilistic forecasts ``for free''.} Chronos-2 provides calibrated probabilistic outputs without retraining or architectural changes, giving urban planners and operators direct access to uncertainty quantification for decision-making (e.g. developing policies associated with a certain degree of confidence). The value of such prediction intervals in transportation has been demonstrated in \cite{Mallick2024-uncertainty, Qian2023-uncertainty}.

\textbf{Potential for fine-tuning.} While zero-shot performance is competitive with or superior to other models across the transportation datasets studied, further gains are possible through fine-tuning on domain-specific data. However, such adaptation can be computationally demanding for large FMs, motivating the use of techniques such as Parameter-Efficient Fine-Tuning (PEFT) \cite{xu2023parameterefficientfinetuningmethodspretrained}. The growing availability of tutorials, open-source codebases, and PEFT frameworks lowers practical barriers and facilitates experimentation.

\textbf{Establishing a new baseline standard.} This work establishes a new benchmark for transportation forecasting. Given their ease of use and strong zero-shot performance, we argue that, moving forward, TS-FMs such as Chronos-2 should serve as default, competitive baselines in future research on transportation forecasting problems (alongside simple historical averages \cite{Rodrigues2023-on-the-importance-of-stationarity} and DL models). We believe that any future domain-specific architectures proposed in the transportation literature would benefit from benchmarking against these generalist models to justify their problem-specific complexities.

\textbf{Risk of homogenization of biases and errors.} At a broader systemic level, as discussed in \cite{bommasani2022opportunitiesrisksfoundationmodels}, using the same FM across many systems means that any systematic bias or zero-shot error in the model can propagate to all downstream applications. In transportation forecasting, this might lead to consistent mispredictions under, e.g., some traffic patterns. For this reason, thorough problem-specific evaluation and monitoring are always necessary to ensure reliable performance.

	
	\bibliographystyle{IEEEtran}
	\bibliography{root} 

@inproceedings{Rodrigues2023-on-the-importance-of-stationarity,
  author    = {Rodrigues, Filipe},
  booktitle = {Proc. IEEE Int. Conf. Intelligent Transportation Systems (ITSC)},
  title     = {On the Importance of Stationarity, Strong Baselines and Benchmarks in Transport Prediction Problems},
  year      = {2023},
  pages     = {4927--4932},
  doi       = {10.1109/ITSC57777.2023.10422030}
}

@article{ansari2024chronoslearninglanguagetime,
  title   = {Chronos: Learning the Language of Time Series},
  author  = {Abdul Fatir Ansari and Lorenzo Stella and Caner Turkmen and Xiyuan Zhang and Pedro Mercado and Huibin Shen and others},
  journal = {Transactions on Machine Learning Research},
  year    = {2024},
  issn    = {2835-8856}
}

@misc{ansari2025chronos2univariateuniversalforecasting,
  title         = {Chronos-2: From Univariate to Universal Forecasting},
  author        = {Abdul Fatir Ansari and Oleksandr Shchur and Jaris K{\"u}ken and Andreas Auer and Boran Han and Pedro Mercado and others},
  year          = {2025},
  eprint        = {2510.15821},
  archivePrefix = {arXiv},
  primaryClass  = {cs.LG}
}

@inproceedings{li2018diffusionconvolutionalrecurrentneural,
  title     = {Diffusion Convolutional Recurrent Neural Network: Data-Driven Traffic Forecasting},
  author    = {Yaguang Li and Rose Yu and Cyrus Shahabi and Yan Liu},
  booktitle = {Int. Conf. Learning Representations (ICLR)},
  year      = {2018}
}

@article{Raftery2007-calibration-sharpness,
  author  = {Gneiting, Tilmann and Balabdaoui, Fadoua and Raftery, Adrian},
  title   = {Probabilistic Forecasts, Calibration and Sharpness},
  journal = {Journal of the Royal Statistical Society: Series B (Statistical Methodology)},
  year    = {2007},
  volume  = {69},
  pages   = {243--268},
  doi     = {10.1111/j.1467-9868.2007.00587.x}
}

@article{li2025urbanev,
  author  = {Li, Han and Qu, Haohao and Tan, Xiaojun and You, Linlin and Zhu, Rui and Fan, Wenqi},
  title   = {{UrbanEV}: An Open Benchmark Dataset for Urban Electric Vehicle Charging Demand Prediction},
  journal = {Scientific Data},
  volume  = {12},
  pages   = {523},
  year    = {2025},
  issn    = {2052-4463},
  doi     = {10.1038/s41597-025-04874-4}
}

@inproceedings{Yu2018-Spatio-Temporal,
  author    = {Yu, Bing and Yin, Haoteng and Zhu, Zhanxing},
  title     = {Spatio-Temporal Graph Convolutional Networks: A Deep Learning Framework for Traffic Forecasting},
  booktitle = {Proc. 27th Int. Joint Conf. Artificial Intelligence (IJCAI)},
  year      = {2018},
  pages     = {3634--3640},
  doi       = {10.24963/ijcai.2018/505}
}

@misc{lee2022ddpgcnmultigraphconvolutionalnetwork,
  title         = {{DDP-GCN}: Multi-Graph Convolutional Network for Spatiotemporal Traffic Forecasting},
  author        = {Kyungeun Lee and Wonjong Rhee},
  year          = {2022},
  eprint        = {1905.12256},
  archivePrefix = {arXiv},
  primaryClass  = {cs.LG}
}

@article{Hewamalage2023ForecastEvaluation,
  author  = {Hewamalage, Hansika and Ackermann, Klaus and Bergmeir, Christoph},
  title   = {Forecast Evaluation for Data Scientists: Common Pitfalls and Best Practices},
  journal = {Data Mining and Knowledge Discovery},
  year    = {2023},
  volume  = {37},
  number  = {2},
  pages   = {788--832},
  doi     = {10.1007/s10618-022-00894-5}
}

@inproceedings{Ye_Sun_Du_Fu_Xiong_2021,
  author    = {Ye, Junchen and Sun, Leilei and Du, Bowen and Fu, Yanjie and Xiong, Hui},
  title     = {Coupled Layer-Wise Graph Convolution for Transportation Demand Prediction},
  booktitle = {Proc. AAAI Conf. Artificial Intelligence},
  year      = {2021},
  volume    = {35},
  number    = {5},
  pages     = {4617--4625},
  doi       = {10.1609/aaai.v35i5.16591}
}

@inproceedings{Choi_Choi_Hwang_Park_2022,
  author    = {Choi, Jeongwhan and Choi, Hwangyong and Hwang, Jeehyun and Park, Noseong},
  title     = {Graph Neural Controlled Differential Equations for Traffic Forecasting},
  booktitle = {Proc. AAAI Conf. Artificial Intelligence},
  year      = {2022},
  volume    = {36},
  number    = {6},
  pages     = {6367--6374},
  doi       = {10.1609/aaai.v36i6.20587}
}

@article{Zhao2020-T-GCN,
  author  = {Zhao, Ling and Song, Yujiao and Zhang, Chao and Liu, Yu and Wang, Pu and Lin, Tao and Deng, Min and Li, Haifeng},
  title   = {{T-GCN}: A Temporal Graph Convolutional Network for Traffic Prediction},
  journal = {IEEE Transactions on Intelligent Transportation Systems},
  year    = {2020},
  volume  = {21},
  number  = {9},
  pages   = {3848--3858},
  doi     = {10.1109/TITS.2019.2935152}
}

@article{Xia2021-3DGCN,
  author  = {Xia, Tong and Lin, Junjie and Li, Yong and Feng, Jie and Hui, Pan and Sun, Funing and Guo, Diansheng and Jin, Depeng},
  title   = {{3DGCN}: 3-Dimensional Dynamic Graph Convolutional Network for Citywide Crowd Flow Prediction},
  journal = {ACM Trans. Knowl. Discov. Data},
  year    = {2021},
  volume  = {15},
  number  = {6},
  pages   = {110:1--110:21},
  issn    = {1556-4681},
  doi     = {10.1145/3451394}
}

@article{Yang2021-real-time-spatiotemporal,
  author  = {Jin-Ming Yang and Zhong-Ren Peng and Lei Lin},
  title   = {Real-Time Spatiotemporal Prediction and Imputation of Traffic Status Based on {LSTM} and Graph {Laplacian} Regularized Matrix Factorization},
  journal = {Transportation Research Part C: Emerging Technologies},
  year    = {2021},
  volume  = {129},
  pages   = {103228},
  issn    = {0968-090X},
  doi     = {10.1016/j.trc.2021.103228}
}

@misc{bommasani2022opportunitiesrisksfoundationmodels,
  title         = {On the Opportunities and Risks of Foundation Models},
  author        = {Rishi Bommasani and Drew A. Hudson and Ehsan Adeli and Russ Altman and others},
  year          = {2022},
  eprint        = {2108.07258},
  archivePrefix = {arXiv},
  primaryClass  = {cs.LG}
}

@inproceedings{Reddy2025_ARIMA,
  title     = {A Methodological Review on Time Series Forecasting by Using {ARIMA}},
  author    = {B. Dilip Kumar Reddy and J. Swami Naik and S. Vinay Kumar and Suresh Kumar and others},
  booktitle = {Proc. Int. Conf. Advanced Materials, Manufacturing and Sustainable Development (ICAMMSD 2024)},
  year      = {2025},
  pages     = {709--719},
  doi       = {10.2991/978-94-6463-662-8_55},
  publisher = {Atlantis Press}
}

@inproceedings{Brown2020_GPT-3,
  author    = {Tom B. Brown and Benjamin Mann and Nick Ryder and Melanie Subbiah and others},
  title     = {Language Models are Few-Shot Learners},
  booktitle = {Advances in Neural Information Processing Systems},
  volume    = {33},
  pages     = {1877--1901},
  year      = {2020}
}

@inproceedings{Devlin2018_BERT,
  author    = {Jacob Devlin and Ming{-}Wei Chang and Kenton Lee and Kristina Toutanova},
  title     = {{BERT}: Pre-Training of Deep Bidirectional Transformers for Language Understanding},
  booktitle = {Proc. 2019 Conf. North American Chapter of the Association for Computational Linguistics: Human Language Technologies (NAACL-HLT)},
  year      = {2019},
  pages     = {4171--4186}
}

@inproceedings{jin2024timellmtimeseriesforecasting,
  title     = {Time-{LLM}: Time Series Forecasting by Reprogramming Large Language Models},
  author    = {Ming Jin and Shiyu Wang and Lintao Ma and Zhixuan Chu and James Y. Zhang and Xiaoming Shi and Pin-Yu Chen and Yuxuan Liang and Yuan-Fang Li and Shirui Pan and Qingsong Wen},
  booktitle = {Int. Conf. Learning Representations (ICLR)},
  year      = {2024}
}

@misc{xue2023promptcastnewpromptbasedlearning,
  title         = {{PromptCast}: A New Prompt-Based Learning Paradigm for Time Series Forecasting},
  author        = {Hao Xue and Flora D. Salim},
  year          = {2023},
  eprint        = {2210.08964},
  archivePrefix = {arXiv},
  primaryClass  = {stat.ME}
}

@inproceedings{das2024googletimeseries,
  title     = {A Decoder-Only Foundation Model for Time-Series Forecasting},
  author    = {Abhimanyu Das and Weihao Kong and Rajat Sen and Yichen Zhou},
  booktitle = {Proc. 41st Int. Conf. Machine Learning (ICML)},
  series    = {Proceedings of Machine Learning Research},
  volume    = {235},
  year      = {2024}
}

@misc{rasul2024lagllamafoundationmodelsprobabilistic,
  title         = {Lag-{Llama}: Towards Foundation Models for Probabilistic Time Series Forecasting},
  author        = {Kashif Rasul and Arjun Ashok and Andrew Robert Williams and others},
  year          = {2024},
  eprint        = {2310.08278},
  archivePrefix = {arXiv},
  primaryClass  = {cs.LG}
}

@article{Raffel2019_T5,
  author  = {Colin Raffel and Noam Shazeer and Adam Roberts and Katherine Lee and Sharan Narang and Michael Matena and Yanqi Zhou and Wei Li and Peter J. Liu},
  title   = {Exploring the Limits of Transfer Learning with a Unified Text-to-Text Transformer},
  journal = {Journal of Machine Learning Research},
  year    = {2020},
  volume  = {21},
  number  = {140},
  pages   = {1--67}
}

@misc{xu2023parameterefficientfinetuningmethodspretrained,
  title         = {Parameter-Efficient Fine-Tuning Methods for Pretrained Language Models: A Critical Review and Assessment},
  author        = {Lingling Xu and Haoran Xie and Si-Zhao Joe Qin and Xiaohui Tao and Fu Lee Wang},
  year          = {2023},
  eprint        = {2312.12148},
  archivePrefix = {arXiv},
  primaryClass  = {cs.CL}
}

@article{Mallick2024-uncertainty,
  author  = {Mallick, Tanwi and Macfarlane, Jane and Balaprakash, Prasanna},
  title   = {Uncertainty Quantification for Traffic Forecasting Using Deep-Ensemble-Based Spatiotemporal Graph Neural Networks},
  journal = {IEEE Transactions on Intelligent Transportation Systems},
  year    = {2024},
  volume  = {25},
  number  = {8},
  pages   = {9141--9152},
  doi     = {10.1109/TITS.2024.3381099}
}

@article{Qian2023-uncertainty,
  author  = {Qian, Weizhu and Zhang, Dalin and Zhao, Yan and Zheng, Kai and Zhou, Xiaofang},
  title   = {Towards a Unified Understanding of Uncertainty Quantification in Traffic Flow Forecasting},
  journal = {IEEE Transactions on Knowledge and Data Engineering},
  year    = {2023},
  volume  = {35},
  number  = {12},
  pages   = {12727--12740},
  doi     = {10.1109/TKDE.2023.3312261}
}
	
\end{document}